\documentclass[sigconf]{acmart}

\AtBeginDocument{%
  \providecommand\BibTeX{{%
    \normalfont B\kern-0.5em{\scshape i\kern-0.25em b}\kern-0.8em\TeX}}}
    
\usepackage{epsfig}
\usepackage{lipsum}
\usepackage{amsmath,graphicx}
\usepackage{tabularx,ragged2e}
\usepackage{subfig}
\usepackage{csquotes}
\usepackage{booktabs}
\usepackage{resizegather}
\usepackage{color,colortbl}
\usepackage{mathtools}
\usepackage{multirow}
\usepackage{balance}

\usepackage{microtype} 
\usepackage{color,colortbl}
\definecolor{Gray}{gray}{0.95}

\usepackage[labelfont=bf,textfont={normal}]{caption}

\copyrightyear{2021}
\acmYear{2021}
\setcopyright{acmcopyright}\acmConference[MM '21]{Proceedings of the 29th ACM International Conference on Multimedia}{October 20--24, 2021}{Virtual Event, China}
\acmBooktitle{Proceedings of the 29th ACM International Conference on Multimedia (MM '21), October 20--24, 2021, Virtual Event, China}
\acmPrice{15.00}
\acmDOI{10.1145/3474085.3475307}
\acmISBN{978-1-4503-8651-7/21/10}

\newcommand{\etal}{\textit{et al.}}
\newcommand{\ie}{\textit{i}.\textit{e}.}
\newcommand{\eg}{\textit{e}.\textit{g}.}
\begin{document}

\fancyhead{}
\title{Skeleton-Contrastive 3D Action Representation Learning}

\author{Fida Mohammad Thoker}
\affiliation{%
  \institution{University of Amsterdam}
 \country{}
  }
\email{f.m.thoker@uva.nl}

\author{Hazel Doughty}
\affiliation{%
  \institution{University of Amsterdam}
 \country{}
  }
\email{hazel.doughty@uva.nl}

\author{Cees G.M. Snoek}
\affiliation{%
  \institution{University of Amsterdam}
  \country{}
  }
\email{cgmsnoek@uva.nl}

\renewcommand{\shortauthors}{Thoker  et al.}

\begin{abstract}
This paper strives for self-supervised learning of a feature space suitable for 
skeleton-based action recognition. 
Our proposal is built upon learning invariances to input skeleton representations and various skeleton augmentations via a noise contrastive estimation. In particular, we propose inter-skeleton contrastive learning, which learns from multiple different input skeleton representations in a cross-contrastive manner. In addition, we contribute several skeleton-specific spatial and temporal augmentations which further encourage the model to learn the spatio-temporal dynamics of skeleton data.  By learning similarities between different skeleton representations as well as augmented views of the same sequence,  the network is encouraged to learn higher-level semantics of the skeleton data than when only using the augmented views. 
Our approach achieves state-of-the-art performance for self-supervised learning from skeleton data on the challenging PKU and NTU datasets with multiple downstream tasks, including  action recognition, action retrieval and semi-supervised learning. Code is available at \textcolor{red}{\href{https://github.com/fmthoker/skeleton-contrast}{https://github.com/fmthoker/skeleton-contrast}}.

\end{abstract}


\maketitle

\section{Introduction}
The goal of this paper is to learn a latent feature space suitable for 3D human action understanding. Different from traditional RGB frames \cite{kuehne2011hmdb,carreira2017quo}, skeleton data consists of 3D coordinates representing the major joints of each person in a video \cite{Shahroudy_2016_NTURGBD,Liu_2019_NTURGBD120,liu2017pku}. It offers a light-weight representation that can be processed faster and in a privacy-preserving manner providing application potential in video surveillance, assisted living, gaming and human-computer interaction. Moreover, when compared to RGB, such a representation is robust to changes in background and appearance \cite{cnn3,rnn3}. However, learning a good feature space for 3D actions requires large amounts of labeled skeleton data \cite{cnn1,cnn4,rnn3,rnn2,stgcn2018aaai,ye2020dynamic,song2020stronger}, which is much harder to obtain than large amounts of labeled RGB video. To address this major shortcoming, we propose a new self-supervised contrastive learning method for 3D skeleton data.
\begin{figure}[t!]
\centering
\includegraphics[width=\linewidth]{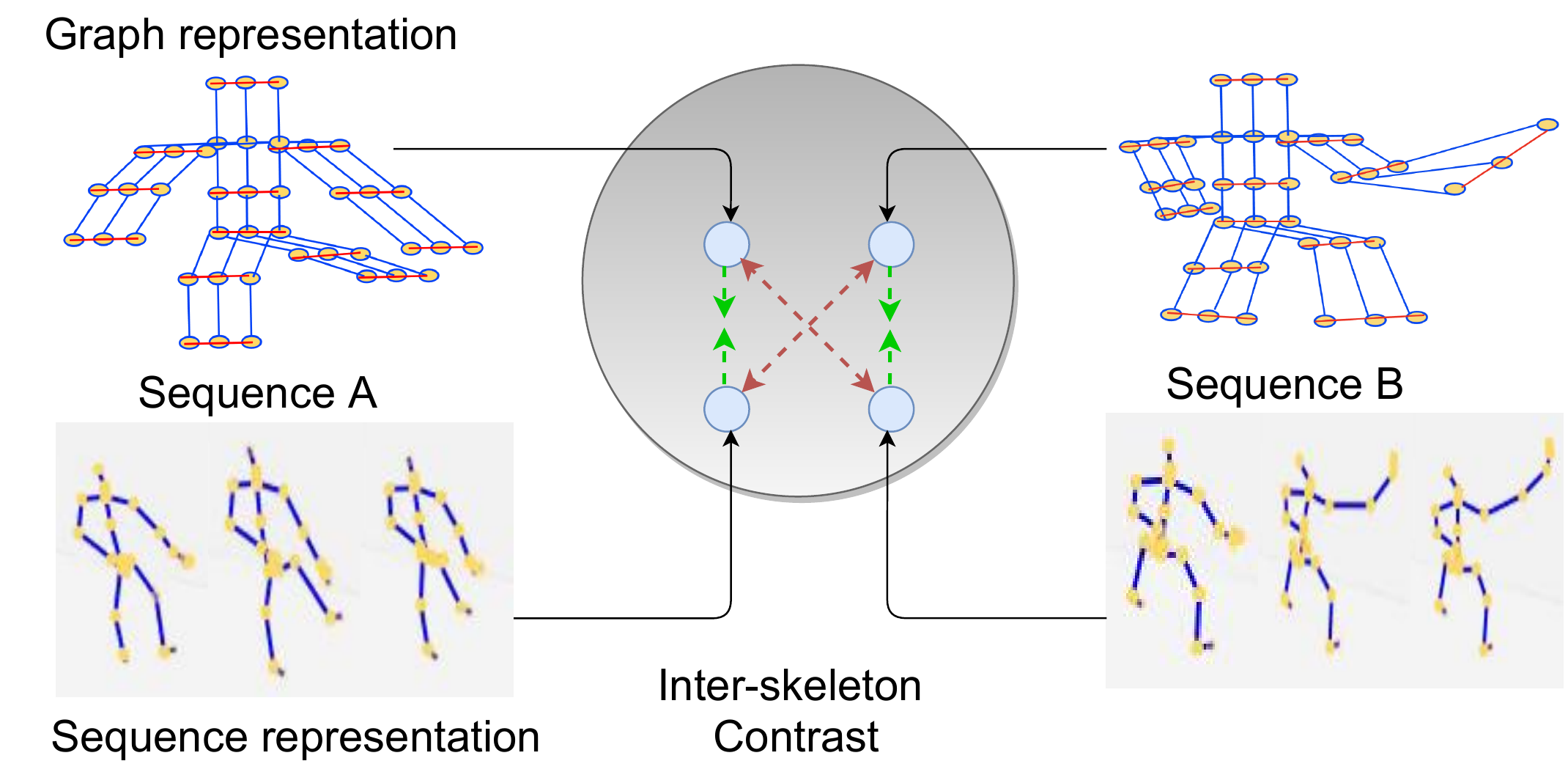}
\caption{\textbf{Inter-skeleton contrast}
learns high-level semantics of skeleton data in a self-supervised fashion. While contrastive methods normally learn invariance to augmentations we additionally learn invariance to the input representation. Different representations of the same sequence are encouraged to be close together in the feature space, while being far away from other sequences.
}
\label{fig:1}
\end{figure}

Several previous works also considered self-supervised learning for 3D skeleton data~\cite{longtan,pc,ms2l,Unsup3DPose}. These works design pretext tasks, such as learning to reconstruct masked input~\cite{longtan} and motion prediction~\cite{ms2l}, which still require the features to represent variations such as the viewpoint and skeleton scale, rather than focusing on higher-level semantic features relevant to downstream tasks.
Instead, we take inspiration from recent self-supervised literature for RGB images, which aims to learn the high-level similarity between augmented forms of the same image and the dissimilarity between these and other images~\cite{infonce, he2019moco, simclr}. At the core of such contrastive learning is the nature of the RGB data, where each sample contains abundant pixel information, allowing for augmentations like spatial-cropping and color-jittering to easily generate subtly different versions of an image without changing its semantic content. However, skeleton sequences are much more sparse than RGB data and the augmentations commonly applied to images would not change the estimated skeleton of a person. Thus, for contrastive learning with skeleton sequences, we need skeleton-specific augmentations to encourage the learned features to encode information relating to spatio-temporal dynamics of the joints. We also want to enrich the input space which can be sampled from, to increase the variety of samples with the same semantic content, and thus increase the difficulty of the contrastive learning task. 

We make three contributions. Our first contribution is to leverage multiple input-representations of the 3D-skeleton sequences. In particular, we propose inter-skeleton contrast to learn from a pair of skeleton-representations in a cross-contrastive fashion, see Figure \ref{fig:1}. This allows us to enrich the sparse input space and focus on the high-level semantics of the skeleton data rather than the nuances of one specific input representation. 
Second, we introduce several skeleton-specific spatial and temporal augmentations for generating positive pairs which encourage the model to focus on the spatio-temporal dynamics of skeleton-based action sequences, ignoring confounding factors such as viewpoint and the exact joint positions. Finally, we provide a comprehensive evaluation of our learned feature space on various challenging downstream tasks, showing considerable improvement over prior methods in all tasks. 

\section{Related Work}
\noindent\textbf{Self-Supervised Learning.}
Self-supervised learning aims to learn feature representations without human annotation, typically by solving \emph{pretext tasks} which exploit the structure of unlabeled data. Previous works have proposed a variety of such tasks for learning image representations, \eg~solving spatial jigsaw puzzles~\cite{jigsaw_image}, rotation prediction~\cite{rotation}, spatial context-prediction~\cite{context}, image inpainting~\cite{inpainting} and colorization \cite{image_color,image_color2}. Similarly, pre-text tasks have been designed for learning video representations, such as spatio-temporal puzzles \cite{puzzle_video}, prediction of frame-order \cite{frame_order}, clip-order \cite{clip_order}, speed \cite{speednet}, future \cite{future} and temporal coherence \cite{coherence}. Such pretext tasks rely on the rich structured nature of RGB data with the hope that by learning to solve these tasks the encoded features will rely on the high-level semantics of the image or video and are thus applicable to the downstream task(s).  Unfortunately, these existing RGB-based pretext tasks are not suited for 3D-skeleton sequences which have a simple structure and are less rich in information. 

Instead of designing specific pretext tasks, recent self-supervised methods rely on instance discrimination and learn the similarity between sample pairs~\cite{infonce,simclr,he2019moco,CMC,pirl}. A noise contrastive loss learns invariances to certain image or video transformation functions, resulting in good feature representations. For example, Chen \etal~\cite{simclr} show that learning invariance to simple image augmentations, such as color jitter, results in highly discriminative features. He \etal~\cite{he2019moco} propose a momentum contrast which is able to utilize a large number of negatives for the noise contrast by storing image features from previous batches in a dynamic queue. In this paper, we rely on contrastive estimation for 3D action representation learning. As existing works use augmentations specific to RGB images, we introduce three skeleton-specific augmentations to generate positive pairs for learning the spatio-temporal dynamics of 3D-skeleton sequences. Furthermore, we propose inter-skeleton contrastive learning which additionally aims to learn invariance to the particular input representation of the 3D-skeleton sequences.

\noindent\textbf{Supervised 3D Action Recognition.} 
Numerous methods for supervised 3D action recognition exist. While earlier methods design handcrafted features~\cite{hand_crafted_1,hand_crafted_2,hand_crafted_3} to model geometric relationships between skeleton joints, recent approaches rely on data-driven deep neural networks. Three skeleton-representations have become popular for deep learning. Sequence-based treats the 3D-skeleton data as a multi-dimensional time-series and models it with a recurrent architecture~\cite{rnn1,rnn2,rnn3,rnn4,Shahroudy_2016_NTURGBD} to learn the temporal dynamics of the joints. Image-based create a pseudo-image representation of the 3D-skeleton data~\cite{cnn1,cnn2,cnn3,cnn4,HCN} which is encoded by  CNN architectures to model the co-occurrence of multiple joints and their motion. Finally, graph-based~\cite{stgcn2018aaai,gcn2,gcn3,gcn4,gcn5,2sagcn2019cvpr,huang2020spatio,peng2020mix} represents the 3D-skeleton data with a graph consisting of spatial and temporal edges. Graph-convolutional architectures then encode the spatio-temporal motion from the human skeleton graph. Although these methods achieve excellent performance, they are all fully supervised and require time-consuming action class annotations. We propose a self-supervised method for 3D-skeleton data that leverages the diversity of the skeleton-representations to learn highly discriminative features from unlabeled data.

\noindent\textbf{Self-Supervised 3D Action Recognition.}
Overcoming the need for large amounts of  annotations has only recently received attention in the 3D action recognition community.
Zheng \etal~\cite{longtan} propose a seq2seq model that learns to reconstruct masked input 3D-skeleton sequences. In particular, a GAN is trained such that the decoder attempts to regenerate the input sequences, while a discriminator measures the quality of the regenerated sequences. Similarly, Nie \etal~\cite{Unsup3DPose} propose a cross-view reconstruction task that relies on a siamese denoising autoencoder to reconstruct the correct version of corrupted and rotated input skeletons.
Su \etal~\cite{pc} also propose a seq2seq model that regenerates input skeleton sequences. To encourage the encoder to learn better latent features, the decoder is weakened by fixing its weights.  

Lin \etal~\cite{ms2l} take a different approach and propose multi-task self-supervised learning for the sequence-based skeleton representation. Their framework solves multiple pretext tasks simultaneously, such as motion prediction and skeleton-jigsaw. 
Si \etal~\cite{si2020adversarial} propose an adversarial self-supervised learning approach that couples the self-supervised learning and the semi-supervised scheme via neighbor relation exploration and adversarial learning. 

Different from all these works, we do not rely exclusively on a sequence-based skeleton-representation and pretext tasks such as input-reconstruction and motion prediction. Instead, we propose to exploit the diversity of skeleton-representations in an inter-contrastive learning regime and design skeleton-specific spatial and temporal augmentations for use in this contrastive method.  
\section{Skeleton-Contrastive Learning}
In this section we present our inter-skeleton contrast approach for self-supervised learning of 3D action features. Contrastive methods aim to learn a good feature space by learning the similarity between augmented views of the same data. Since augmentations in existing contrastive learning works are primarily designed for RGB images \cite{simclr} they are not suitable for the skeleton data that is considered in this work. Therefore, we first propose several skeleton-specific augmentation functions in Section~\ref{section-3.1}. These augmentations enable us to apply existing contrastive learning methods, such as MoCo~\cite{he2019moco}, to skeleton data. We describe this is Section~\ref{section-3.2}.

However, contrastive learning can be vulnerable to shortcuts, where simple features, irrelevant to the downstream task, may be enough to identify the different augmented views of the same data. For instance, Chen \etal~\cite{simclr} show that color distributions can be a shortcut to identify different crops from the same image.
To avoid such shortcuts and make the contrastive learning task more difficult, we additionally contrast pairs of different input  skeleton representations with each other. We call this \textit{inter-skeleton contrastive learning} and detail our approach in Section~\ref{section:-3.3}. 

\begin{figure}[t!]
\centering
\includegraphics[width=\linewidth]{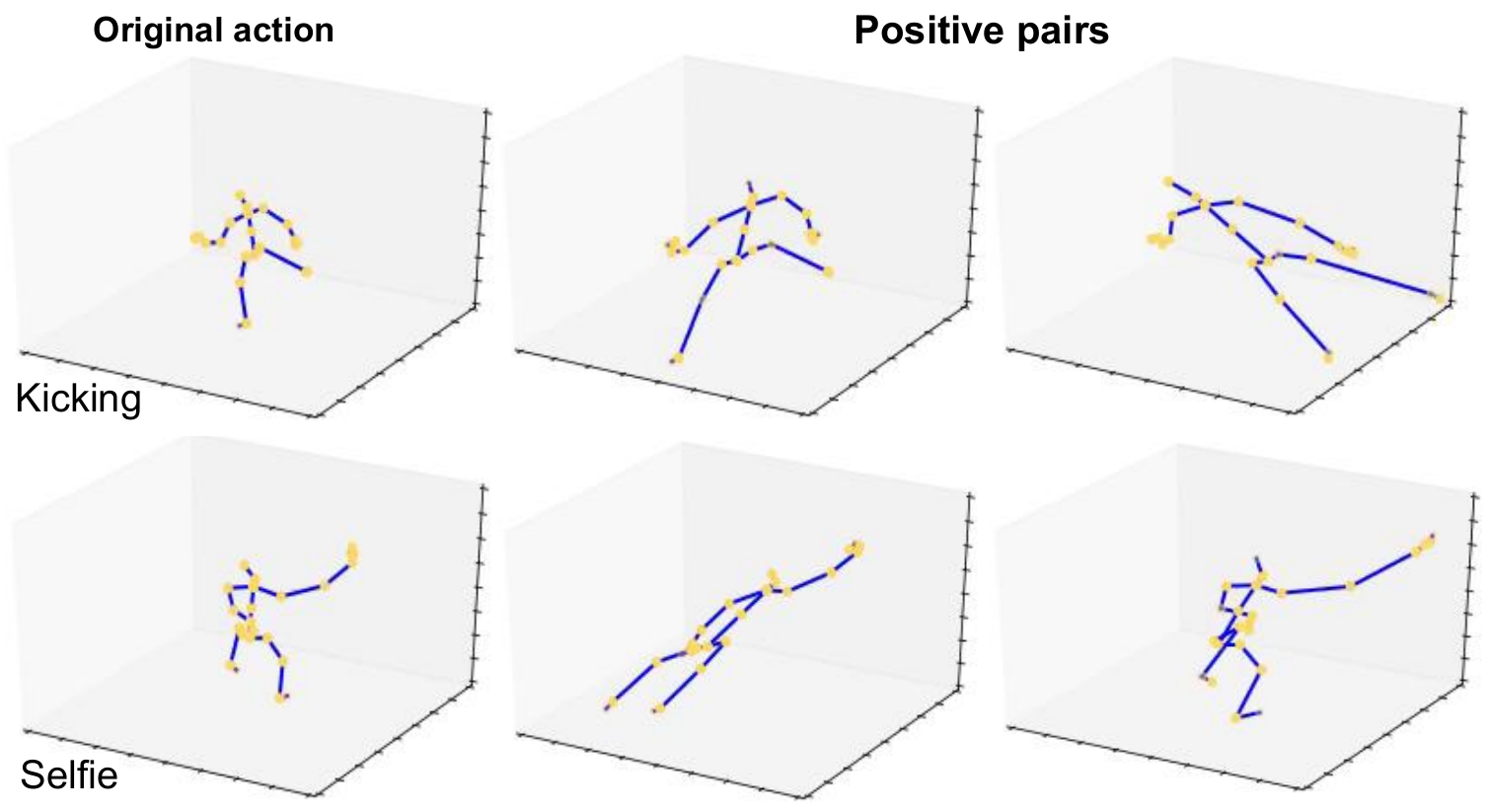}
\caption{\textbf{Spatial pose augmentation} examples.  A shear operation is applied to the original action so that the augmented pairs differ in viewpoint and camera distance. 
}
\label{fig:2}
\end{figure}
 
\subsection{Skeleton Augmentations} 
\label{section-3.1}
The goal of contrastive learning is to learn the semantic similarity between items in a dataset without labels. This is usually done by learning the similarity of two augmented views (positive pairs) of a sample $X$. A data augmentation function $D$, composed of a single or multiple transformations, creates the augmented views. Hence, the network learns features for $X$, which are invariant to the transformations in $D$. The nature of the data $X$ and the downstream task determines the appropriate invariances that the learned features should possess.
In our case, $X$ is a 3D-skeleton sequence, where each sequence represents a particular spatial configuration of human joints and its motion over a short period of time. Thus, to learn useful representations for 3D-skeleton data, the commonly used RGB augmentations, such as color-distortion and Gaussian blurring~\cite{simclr}, are not suitable. Instead, we need to learn invariances to transformations that encode the spatial and temporal dynamics of 3D skeleton action sequences. We introduce multiple spatial and temporal skeleton augmentation techniques to generate positive pairs for 3D-skeleton action sequences: \textit{Pose Augmentation}, \textit{Joint Jittering} and \textit{Temporal Crop-Resize}. We then combine these to create our final spatio-temporal skeleton augmentation. Let us assume each raw action sequence $X \in R^{T{\times} J{\times}3}$ consists of 3D coordinates of $J$ body joints in $T$ consecutive video frames. We define our individual augmentations $D$ based on $X$. 

\begin{figure}[t!]
\centering
\includegraphics[width=\linewidth]{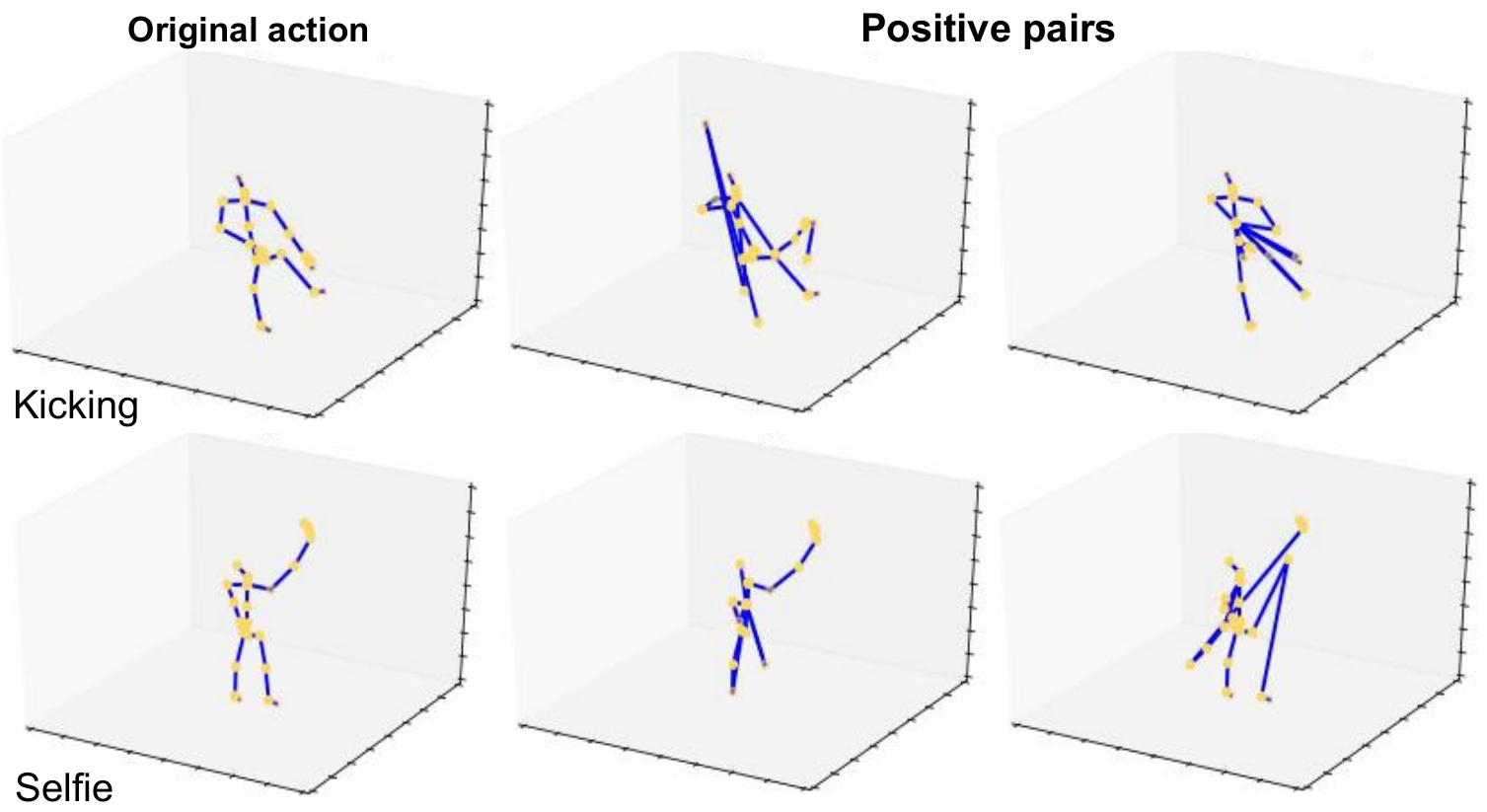}
\caption{\textbf{Spatial joint jittering} examples.  The augmented pairs contain a subset of common joint connections while other joint connections are randomly moved to an irregular position.
}
\label{fig:3}
\end{figure}

\subsubsection{\textbf{Spatial Skeleton Augmentations}}
To apply our learned feature space to downstream tasks such as 3D action recognition, we require the feature encodings to rely on more discriminatory spatial semantics like joint configurations, while being invariant to factors such as viewpoint, camera distance, skeleton scale and joint perturbations. Existing augmentations for RGB images would not achieve this, thus we propose two new skeleton-specific spatial augmentations: pose augmentation and joint jittering. These can be applied to each of the $T$ skeletons in the sequence $X$ so a contrastive learning framework can learn invariance to these augmentations.

\noindent\textbf{Pose Augmentation.} 
With this transformation, we aim to create positive pairs which differ in viewpoint and distance to the camera, while retaining the same pose from the original sequence. To achieve this, we apply a 3D shear on the action sequence $X$:
\begin{equation}
D_{{\textit{Spatial}}_1}(X) = 
 X \cdot \begin{bmatrix}
   1& r_{01} & r_{02} \\
    r_{10} & 1& r_{12} \\
    r_{20} &  r_{21} & 1 \\
   \end{bmatrix}, 
\end{equation}
where the elements of the augmentation matrix are randomly drawn from a uniform distribution $[-1,1]$.
Figure~\ref{fig:2} shows several examples. By applying the same shearing operation to each joint of the skeleton at each time-step in the sequence we are able to simulate changes in camera viewpoint and distance between the subject and camera. Therefore, a contrastive network which learns invariance to this transformation is forced to learn more discriminatory pose semantics of the positive pairs and ignore redundant information such as the viewpoint and proximity to the camera.

\noindent\textbf{Joint Jittering.} 
We also want a contrastive method to be invariant to noise in the estimated skeleton. Therefore we propose joint jittering to create positive pairs where some of the joint connections in $X$ are randomly perturbed. We select $j$ of the $J$ joints at random and move these joints to irregular positions, while keeping other joints in their original position. The transformation is defined as:
\begin{equation}
D_{{\textit{Spatial}}_2}(X) = 
 X[:,j]\cdot\begin{bmatrix}
   r_{00}& r_{01} & r_{02} \\
    r_{10} & r_{11} & r_{12} \\
    r_{20} &  r_{21} & r_{22} \\
   \end{bmatrix} , 
\end{equation}
where $j$ is a subset of the joints such that $|j| < J$, and the elements of the jitter matrix are randomly drawn from a uniform distribution $[-1,1]$. The same jitter matrix is applied to each joint in $j$ at each time-step $T$. Examples are shown in Figure \ref{fig:3}. 
To learn invariance to such transformations, the contrastive task is encouraged to rely on the spatio-temporal semantics of the common joint connections and ignore the noise from the irregular joint connections. 

\subsubsection{\textbf{Temporal Skeleton Augmentation}}
Besides the spatial perturbations, a good 3D skeleton feature space should also be robust to temporal modifications of the skeleton sequences, such as the speed of an action and changes to the temporal bounds of the sequence. To this end, we propose temporal crop-resize.

\begin{figure}[t!]
 \centering
 \includegraphics[width=\linewidth,]{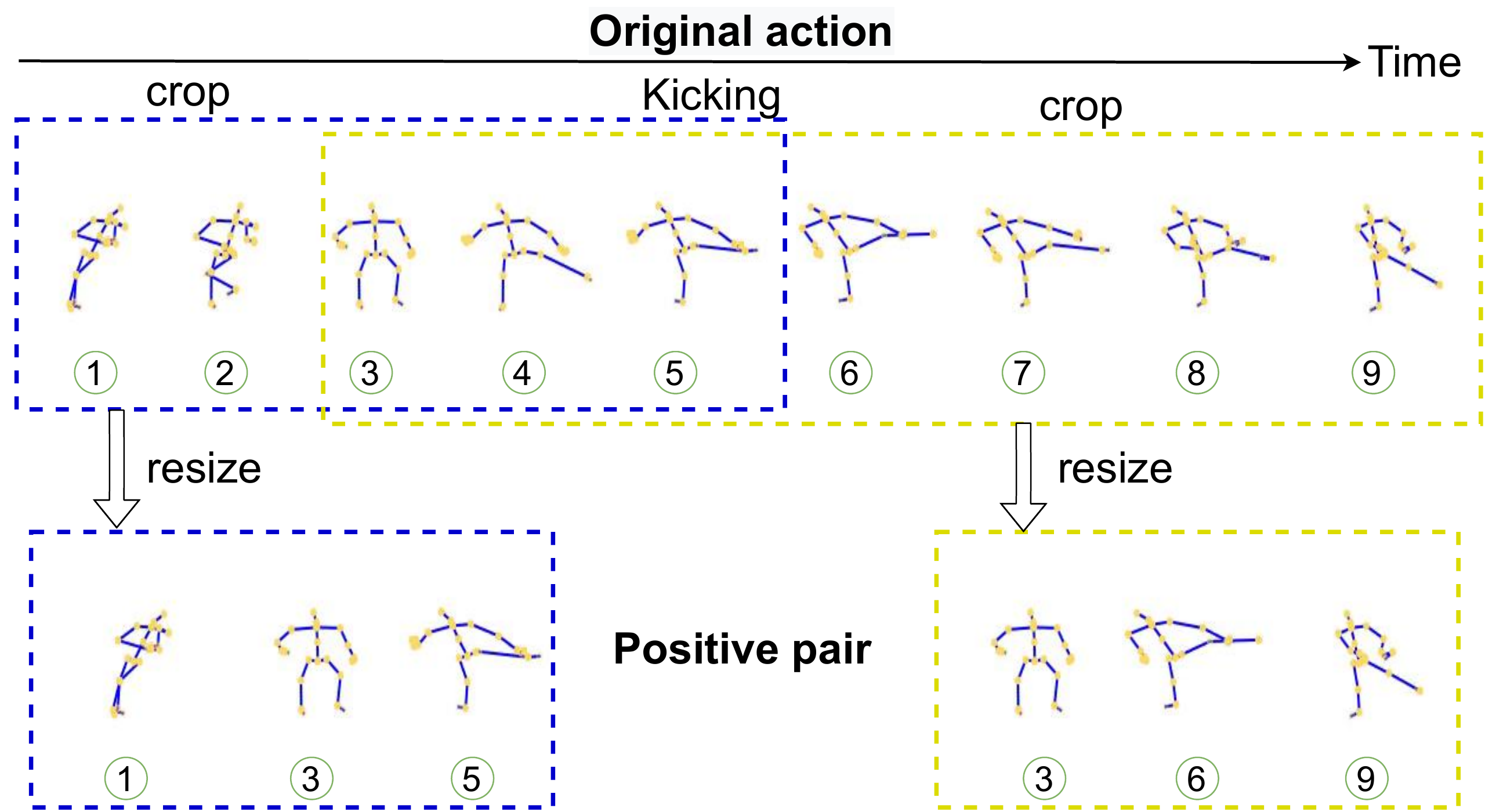}
\caption{\textbf{Temporal crop-resize}. The augmented views start at different time steps and sample different temporal periods (blue and yellow boxes). Each crop is re-sampled to a fixed size, effectively altering its speed depending on the length of the temporal crop. 
}
\label{fig:4}
\end{figure}

\noindent\textbf{Temporal Crop-Resize.}  In this transformation, we create positive pairs with varying speed and varying starting and ending points. We sample different parts of the action sequence $X$ via a random crop and resize this crop over the temporal dimension $T$:
\begin{equation}
\label{eq:cropresize}
D_{\textit{Temporal}}(X) = 
 \mathrm{Interpolate}(X[L_{\textit{start}}: L_{\textit{start}} + TL_{\textit{ratio}}]). 
\end{equation}
The length ratio $L_\textit{ratio}$ is first randomly sampled from distribution $[l_{min}, 1.0]$, followed by randomly selecting a starting frame $L_\textit{start}$ between $(0, T - TL_\textit{ratio})$. The sub-sequence $X[L_\textit{start}: L_\textit{start}+ TL_\textit{ratio}]$ is then re-sampled to a fixed length.  This re-sampling causes the temporal crop-resize to also alter the speed of a sequence as well as its start and end times; a shorter sub-sequence will effectively have a slower speed once re-sampled.
Figure~\ref{fig:4} shows examples of this transformation. By including this augmentation the contrastive task is forced to focus on the commonalities of the joint motion dynamics over the sampled temporal periods and be robust to changes in the exact start, end and speed of an action.

\subsubsection{\textbf{Spatio-Temporal Skeleton Augmentations.}}
To learn the spatio-temporal dynamics of the skeleton sequences, we propose to combine the above spatial and temporal transformations into a single augmentation function. 
Such composition results in strong positive pairs which vary in both spatial and temporal dynamics locally, while retaining the high-level semantics of the original action sequence.
In particular, we first apply the temporal crop-resize augmentation $D_{\textit{Temporal}}$ on the original action sequence $X$ followed by a spatial augmentation $D_{{\textit{Spatial}}_{i}}$ to the resulting sequence:
\begin{equation}
D_{\textit{Spatio-Temporal}} (X) =  D_{{\textit{Spatial}}_i}  (D_{\textit{Temporal}}(X)) 
\end{equation}
Here, $i$ can either be fixed to the pose augmentation or the joint jitter or randomized to select either of the spatial augmentations.
 As we will show in the experiments, learning invariance to spatio-temporal transformations produces a better 3D action feature space and randomizing the composition further improves the result. 

\begin{figure*}[t!]
\centering
\includegraphics[width=\linewidth]{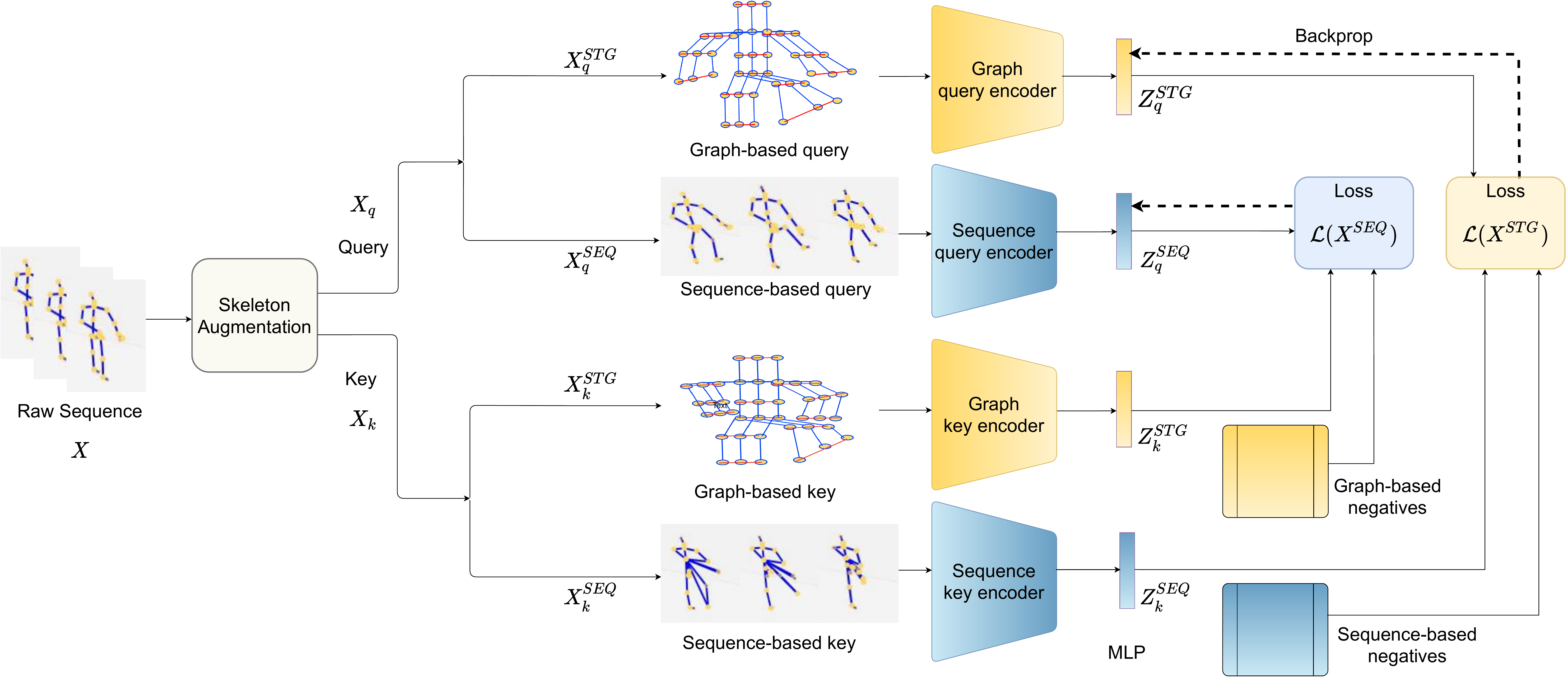}
\caption{\textbf{Inter-skeleton contrast}. We learn invariances to input skeleton representations, as well data augmentations, in a cross-contrastive manner. We first augment the input sequence into two different views called the query and key using our proposed spatio-temporal augmentations. Each of these views is then represented with two different input skeleton-representations, here graph-based and sequence-based. We encourage the embedding for the graph-based query to be similar to the embedding of the sequence-based key while being dissimilar to the current set of sequence-based negatives. The same applies for the sequence-based query and graph-based key and negatives.
}
\label{fig:5}
\end{figure*}

\subsection{Intra-Skeleton Contrast} 
\label{section-3.2}
Before describing our proposed inter-skeleton method, we first describe how the above augmentations can be incorporated into an existing contrastive method, such as MoCo~\cite{he2019moco}, with a single input skeleton-representation. We call this intra-skeleton contrastive learning. Each raw action sequence $X \in R^{T{\times} J{\times}3}$ is first augmented into two different views $X_q$ and $X_k$ (called query and key) via a data augmentation function $D$. Both views of the skeleton data are then instantiated into the same skeleton-representation, be it image-based or sequence-based or graph-based. A contrastive method such as MoCo uses two encoders, one for the query and one for the key.  
We refer to the query encoder as $f_q$ and the key encoder as $f_k$. Let $(Z_q,Z_{k})=(f_q(X_q),f_k(X_k))$ be output embeddings of the encoders for the input query-key pair. 
We then train the contrastive network using the noise contrastive estimation loss InfoNCE \cite{infonce}: 

\begin{equation}
\label{loss:intra}
\mathcal{L}(X) = -\log \frac{\displaystyle \mathrm{exp}(Z_{q}\cdot Z_{k}/\tau)}
{ \displaystyle\mathrm{exp}(Z_{q}\cdot Z_{k}/\tau)   + \displaystyle\sum_{Z_n \sim \mathcal{N}} \mathrm{exp}(Z_{q}\cdot Z_{n}/\tau)},
\end{equation}
where $\tau$ is a temperature softening hyper-parameter and $\mathcal{N}$ is the current set of negatives that  are stored in a dynamic queue via previous states of the key encoder $f_k$ as in ~\cite{he2019moco}. Only the query encoder is actively trained using Equation \eqref{loss:intra} and the key encoder is updated as a moving average of the query encoder. This trains the framework to learn  3D action features which are invariant to the transformations in $D$ for the chosen skeleton-representation. 

\subsection{Inter-Skeleton Contrast}
\label{section:-3.3}
Up to this point, our method, like previous contrastive learning approaches \cite{simclr,he2019moco,pirl}, learns the similarity between different augmented forms of the same input.
We now extend contrastive learning for 3D skeleton data beyond these augmentations and propose inter-skeleton contrast which aims to learn invariance to the \textit{input representation} of the skeleton sequence. 
Three 3D-skeleton representations are common: \textit{image-based} as a $T \times J$ pseudo-image where the 3D coordinates of each joint are the image channels, \textit{sequence-based} as a multi-dimensional time series, or \textit{graph-based} as a spatio-temporal graph. Each requires a different network architecture and encodes different characteristics of the sequence. For example, RNNs treat skeleton sequences as a time series and explicitly model the temporal evolution of joints, while GCNs treat sequences as a graph with both spatial and temporal edges and thus explicitly encode human pose as well as each joint's temporal motion. 
While the action depicted by the skeleton sequence is the same, the way the input sequence is represented and encoded is different.
To learn invariance to the input representation the contrastive framework has to learn the similarities between the characteristics of these different representations as well as our data augmentations which will result in more discriminative features.  

The overall network is depicted in Figure~\ref{fig:5}. The raw skeleton sequence is first augmented into two views as in Section~\ref{section-3.2}. Each view is then represented in two ways, in this case with a graph-based representation and a sequence-based representation. 
We refer to the different representations of the raw action sequence $X$ as $X^{\textit{IMG}}$ for image-based, $X^{\textit{SEQ}}$ for seq-based  and $X^{\textit{STG}}$ for graph-based. 
For the rest of this section we will take the example of the pair $X^{\textit{SEQ}}$ and $X^{\textit{STG}}$ as displayed in Figure~\ref{fig:5}. 
We adapt our model to contrast the different input representations by using a pair of momentum contrastive models together, one for each input-representation $X^{\textit{SEQ}}$ and $X^{\textit{STG}}$. In particular, the model now consists of two query encoders $f^{\textit{SEQ}}_q$ and $f^{\textit{STG}}_q$ and two key encoders $f^{\textit{SEQ}}_k$ and $f^{\textit{STG}}_k$. A query-key pair $(X_q,X_{k})$ is obtained by augmenting a raw action sequence $X$ with $D$ as before. We instantiate two different skeleton-representation pairs ($X_{q}^{\textit{SEQ}},X_{k}^{\textit{SEQ}})$ and ($X_{q}^{\textit{STG}},X_{k}^{\textit{STG}})$. 
 Then, for the query in each input representation, we generate the positives and negatives from the key encoder of the \textit{other} input representation and vice versa. 
 The encoders $(f^{\textit{SEQ}}_q,f^{\textit{STG}}_q)$ are trained jointly using a cross-contrastive loss function:
\begin{equation}
\label{loss:inter}
\mathcal{L}(X^{\textit{SEQ}},X^{\textit{STG}}) =  \mathcal{L}(X^{\textit{SEQ}}) + \mathcal{L}(X^{\textit{STG}}),
\end{equation}
\begin{equation}
\mathcal{L}(X^{\textit{SEQ}}) = -\log \frac{\displaystyle{\mathrm{exp}( Z_{q}^{\textit{SEQ}}\cdot Z_{k}^{\textit{STG}}/\tau) }}
{ \displaystyle\mathrm{exp}(Z_{q}^{\textit{SEQ}}\cdot Z_{k}^{\textit{STG}}/\tau)  + \displaystyle\sum_{Z_n \sim \mathcal{N}^{\textit{STG}}} \mathrm{exp}(Z_{q}^{\textit{SEQ}}\cdot Z_{n}^{\textit{STG}}/\tau)},
\end{equation}
\begin{equation}
\mathcal{L}(X^{\textit{STG}}) =  -\log \frac{\displaystyle{\mathrm{exp}(Z_{q}^{\textit{STG}}\cdot Z_{k}^{\textit{SEQ}}/\tau) }}
{ \displaystyle{\mathrm{exp}(Z_{q}^{\textit{STG}}\cdot Z_{k}^{\textit{SEQ}}/\tau) }  + \displaystyle\sum_{Z_n \sim \mathcal{N}^{\textit{SEQ}}} \mathrm{exp}(Z_{q}^{\textit{STG}}\cdot Z_{n}^{\textit{SEQ}}/\tau)},
\end{equation}
where $Z_{q}^{\textit{SEQ}}{=}f_{q}^{\textit{SEQ}}(X_{q}^{SEQ})$ is the embedding of the sequence-based query and $\mathcal{N}^{\textit{SEQ}}$ is the current set of negative sequence-based embeddings. These are defined similarly for the other representations and augmentations of $X$.
This formulation serves two purposes. First the input space of the contrastive task is enriched to learn from multiple representations of the same sequence, in addition to the multiple `views' the data augmentation $D$ provides. Second, different from Equation \eqref{loss:intra}, the cross-contrastive loss \ie~Equation \eqref{loss:inter} forces the framework to rely on mutual information between the embeddings of the two skeleton representations. Thus the contrastive framework is encouraged to focus on higher-level semantics and avoid resorting to shortcut solutions to identify the similarity between query-key pairs.
\section{Experiments}
\begin{table*}[t!] 
 \centering
 \resizebox{0.75\linewidth}{!}{
 \begin{tabular}{lcccccc}
 \toprule
  & \multicolumn{2}{c} {\textbf{NTU RGB+D 60}} & \multicolumn{2}{c} {\textbf{NTU RGB+D 120}} & {\textbf{PKU-MMD I}} & {\textbf{PKU-MMD II}}  \\ \cmidrule(lr){2-3} \cmidrule(lr){4-5} \cmidrule(lr){6-6}  \cmidrule(lr){7-7}   
   &  x-view & x-sub & x-setup & x-sub & x-sub &  x-sub\\
  \midrule
 Zheng \etal~\cite{longtan} &  56.4  & 52.1 & 39.7 & 35.6 &  68.7 & 26.5    \\
 Lin \etal~\cite{ms2l}      &  --     & 52.5 & -- & -- & 64.8 & 27.6     \\
 Su \etal~\cite{pc}         &  59.3  & 56.1 & 44.1 & 41.1  & 59.9 & 25.5      \\
 Nie \etal~\cite{Unsup3DPose}   & 79.7 &  --  & -- & -- & --  & --        \\
\textbf{\textit{This paper}}  &  \textbf{85.2} & \textbf{76.3} &  \textbf{67.9}  & \textbf{67.1} & \textbf{80.9} & \textbf{36.0} \\
 \bottomrule

 \end{tabular}
 }
 \caption{\textbf{3D action recognition.} Our method learns better 3D-action features from unlabeled data than alternatives, no matter the dataset or evaluation protocol. All results of Zheng \etal~and Su \etal~obtained with code provided by Su~\etal }
 \label{table:1}
 \vspace{-1em}
\end{table*}
We first describe the datasets and implementation details. We then demonstrate the effectiveness of our contrastive learning approach on several 3D action understanding downstream tasks. Finally, we ablate the effects of our proposed skeleton augmentations and inter-skeleton contrast.
\subsection{Datasets and Evaluation}

\textbf{NTU RGB+D 60}~\cite{Shahroudy_2016_NTURGBD}. This is the most commonly used dataset for 3D action recognition. All actions are captured in indoor scenes with three cameras concurrently. The dataset contains 40 different subjects and 60 action classes.  Each action sequence is performed by an individual or pair of actors with each actor represented by the 3D coordinates of 25 skeleton joints. The dataset consists of 56,880 video samples and is evaluated under the two standard protocols as suggested by \cite{Shahroudy_2016_NTURGBD}. The first is \textit{cross-view}, where samples from two angles ($0^\circ,45^\circ $) are used for training (37,920 samples) and a third angle ($-45^\circ$) is used for testing (18,960 samples). The second is \textit{cross-subject}, where the actors in the training and testing sets are different, with 40,320 training and 16,560 testing samples.

\noindent\textbf{NTU RGB+D 120}~\cite{Liu_2019_NTURGBD120}. This is an extension to NTU RGB-D 60 and is currently the largest benchmark for 3D action recognition with 114,480 samples over 120 action classes. Actions are captured with 106 subjects in a multi-view setting using 32 different setups (varying camera distances and background).
Each action sample has 1 or 2 subjects, and each is represented by 25 3D-skeleton joints. The dataset is challenging due to the variation in subject, background, viewpoint and fine-grained actions captured. For evaluation, two recommended protocols \cite{Liu_2019_NTURGBD120} are used: \textit{cross-setup}, where even-numbered setups are used for training (54,471 samples) and odd-numbered setups are used for testing (59,477 samples), and again \textit{cross-subject}, with 63,026 training and 50,922 testing samples.

\noindent\textbf{PKU-MMD}~\cite{liu2017pku}. This dataset was originally proposed for action detection but has also been used for action recognition~\cite{ms2l}. It contains 52 human action classes. Each action is represented by the 3D coordinates of the 25 joints of each actor involved in the action. The dataset consists of two parts: \textbf{\textit{PKU-MMD I}} and \textbf{\textit{PKU-MMD II}}, with almost 20,000 and 7,000 action instances. Both parts are challenging for action recognition, as the number of action classes is large while the training sets are relatively small, however PKU-MMD II is more challenging due to the large view variation causing more skeleton noise.  
We split both sets into a training and a testing set using the recommended \textit{cross-subject} protocol \cite{liu2017pku}. The training sets of PKU-MMD I \& II contain 18,841 and 5,332 samples, while the testing sets contain 2,704 and 1,613 samples.

\noindent\textbf{Evaluation Criteria.} For all datasets, protocols and downstream tasks we report the top-1 accuracy.

\subsection{Implementation Details}
\label{section:implementation}

\textbf{Network Architectures.} 
We instantiate the encoder pairs $(f_{q},f_{k})$ based on the skeleton-representations used. For the sequence-based encoder $f^{\textit{SEQ}}$ we rely on a 3-Layer BI-GRU with $H{=}$1024 units per layer~\cite{pc}. For the image representation encoder $f^{\textit{IMG}}$, we adopt the CNN based Hierarchical Co-occurrence Network (HCN)~\cite{HCN}. For the graph representation encoder $f^{\textit{STG}}$, a joint based graph-convolutional network A-GCN \cite{2sagcn2019cvpr} is used. We represent each skeleton sequence $X$ as two people, with the second actor being all zeros for single actor actions. The augmented forms of the raw skeleton sequence $X$ ($X_q$ and $X_k$) have temporal length 64. Unless mentioned otherwise we use $|j| = 15$ for the joint jitter augmentation and $l_{min}=0.1$ for the temporal crop-resize augmentation.

\noindent\textbf{Self-Supervised Pretraining.}
Our inter-skeleton contrastive network is based on MOCO~\cite{he2019moco} and is trained on the training data without any labels. A projection head (an MLP) is appended to each encoder to produce embeddings of a fixed size of 128. The embeddings are L2-normalized before computing the contrastive loss. We train the whole network  with a temperature value of $\tau{=}$0.07, an SGD optimizer, a learning rate of 0.01 and a weight decay of 0.0001. 
For NTU RGB+D 60 \& 120, the size of the set negatives $\mathcal{N}$ is 16,384 and the model is pre-trained for a total of 450 epochs. For PKU-MMD I \& II, the size of $\mathcal{N}$ is set to 8,192 and 2,048, and we pre-train for 600 epochs. The training and evaluation details of the downstream tasks are discussed in the supplementary material.
\subsection{Downstream Tasks}
\label{section:downstream}

\begin{table*}[t!]
  \resizebox{\linewidth}{!}{
 \centering
 \begin{tabular}{lllllllllll}
 \toprule
  & \multicolumn{8}{c}{\textbf{NTU RGB+D 60 }} 
  &\multicolumn{2}{c}{\textbf{PKU-MMD I }}  \\
  \cmidrule(lr){2-9} \cmidrule(lr){10-11}
  & \multicolumn{4}{c}{\textbf{x-view}} & \multicolumn{4}{c}{\textbf{x-sub}} &\multicolumn{2}{c}{\textbf{x-sub}}  \\
  \cmidrule(lr){2-5} \cmidrule(lr){6-9} \cmidrule(lr){10-11} 
  &   (1\%) &  (5\%) & (10\%) &  (20\%) & (1\%) &  (5\%) & (10\%) &  (20\%) & (1\%) &  (10\%) \\
  \midrule
 Zheng \etal ~\cite{longtan}    &  - & - & - & - & 35.2 & - & 62.0 & - & 34.4    &69.5  \\
 Lin \etal~\cite{ms2l}   &  - & - & - & - & 33.1 & - & 65.1 & - & 36.4    &70.3  \\
 Si \etal~\cite{si2020adversarial} &  - & 63.6 & 69.8 & 74.7 & - & 57.3 & 64.3 & 68.0 & - & -  \\
\textbf{\textit{This paper}} (supervised only)   &  {21.7} \scriptsize{$\pm$1.0}    & {47.6} \scriptsize{$\pm$1.0}    & {59.8} \scriptsize{$\pm$0.5}    & 69.1 \scriptsize{$\pm$0.5} & 17.6 \scriptsize{$\pm$0.5} & 42.8 \scriptsize{$\pm$0.5} & 51.6 \scriptsize{$\pm$1.0}   & 59.5 \scriptsize{$\pm$1.0}  & 22.5 \scriptsize{$\pm$1.0} & 55.4 \scriptsize{$\pm$1.0}   \\

\textbf{\textit{This paper}}   &  \textbf{38.1} \scriptsize{$\pm$1.0}   &  \textbf{65.7} \scriptsize{$\pm$0.5}   &\textbf{72.5} \scriptsize{$\pm$0.4}   &\textbf{78.2} \scriptsize{$\pm$0.3}   &\textbf{35.7} \scriptsize{$\pm$0.5}   &\textbf{59.6} \scriptsize{$\pm$0.5}   &\textbf{65.9} \scriptsize{$\pm$1.0}   &\textbf{70.8} \scriptsize{$\pm$1.0}   &\textbf{37.7} \scriptsize{$\pm$1.0}   &\textbf{72.1} \scriptsize{$\pm$1.0}   \\
 \bottomrule
 \end{tabular}
 }
 \caption{\textbf{Semi-supervised 3D action recognition.}  We report average accuracy of five runs with random subsets of labeled samples. Pre-training with our inter-skeleton shows improvement over prior semi-supervised works 
 as well as training only with the labeled subset. 
} \label{table:2}
\vspace{-1em}
\end{table*}

\begin{table}[t!] 
 \centering
 \resizebox{\linewidth}{!}{
 \begin{tabular}{lcccc}
 \toprule
  & \multicolumn{2}{c}{\textbf{NTU RGB+D 60}} & \multicolumn{2}{c}{\textbf{NTU RGB+D 120}} \\
  \cmidrule(lr){2-3} \cmidrule(lr){4-5} 
   &  x-view & x-sub & x-setup & x-sub \\  
 \midrule
 Zheng \etal~\cite{longtan} &  48.1  & 39.1 & 35.5 & 31.5 \\
 Su \etal~\cite{pc}         &  76.3  & 50.7 & 41.8 & 39.5 \\
 \textbf{\textit{This paper}} &  \textbf{82.6}  & \textbf{62.5} & \textbf{52.3}  & \textbf{50.6} \\
 \bottomrule
 \end{tabular}
 }
 \caption{
 \textbf{3D action retrieval.}{ Results for Zheng~\etal ~and Su~\etal ~in~\cite{pc} obtained with code provided by Su \etal~Our method learns best features for retrieval than prior self-supervised methods. 
 }}
\label{table:3}
 \vspace{-1em}
\end{table}

In this section, we evaluate the 3D action features learned by our inter-skeleton contrast for various downstream tasks in comparison with the respective state-of-the-art in self-supervised learning.
For a fair comparison we follow the setups of prior works and only train and evaluate downstream tasks with the sequence-based input representation $X^{SEQ}$.  In particular, we pre-train our inter-skeleton contrast network with $X^{SEQ}$ and $X^{STG}$ skeleton representations as this gives the best result (see Section~\ref{sec:ablation}) and evaluate only the sequence-based query encoder $f_q^{SEQ}$. We also show some qualitative results in the supplementary material.

\noindent\textbf{3D Action Recognition.}
  We compare our method to prior works in self-supervised learning for skeleton data by training a linear classifier on top of the frozen features from our inter-skeleton contrast. We compare with the proposed methods of Zheng \etal~\cite{longtan}, Su \etal~\cite{pc} and Nie \etal~\cite{Unsup3DPose}, all of which use reconstruction of the skeleton sequence as a pretext task. We also compare to the multi-task self-supervised method by Lin \etal~\cite{ms2l}, which uses skeleton-jigsaw and motion prediction as auxiliary tasks.
 
 We present results on the NTU RGB+D 60, NTU-120 and PKU-MMD (I and II) datasets in Table \ref{table:1}. It is evident our inter-skeleton contrast outperforms all methods by a considerable margin on each benchmark. We conclude the self-supervised feature space learned by our method is state-of-the-art for 3D action recognition.

\noindent\textbf{3D Action Retrieval.} We follow the setup introduced by Su \etal~\cite{pc}. We apply the $k$NN classifier ($k{=}1$) to the pre-trained features of the training set to assign classes. We match each test sample to the most similar training class using cosine similarity. Besides comparison with Su \etal~\cite{pc}, we also compare with Zheng \etal~\cite{longtan}, using numbers and code provided by Su \etal~We present results for NTU RGB+D 60 and NTU RGB+D 120 in Table \ref{table:3}. For both datasets, our method outperforms the alternatives, especially for the more challenging cross-subject and cross-setup protocols. Both \cite{longtan,pc} rely on an input reconstruction pretext-task for learning their feature space, which easily captures varying viewpoints. However, with a simple reconstruction, it is difficult to capture variation with respect to subjects and setups as our inter-skeleton contrast can.

\noindent\textbf{Semi-Supervised 3D Action Recognition.}
In semi-supervised setting, a network utilizes both labeled and unlabeled data during the training process. 
Following prior work for semi-supervised learning in 3D action recognition, we first train our encoder on our unsupervised inter-skeleton contrastive learning task. Then, we fine-tune the final classification layer and the pre-trained encoder together using a portion of the data labeled with the action class. 
 Again, we compare with Zheng \etal~\cite{longtan} and Lin \etal~\cite{ms2l} as well as the method of Si \etal~\cite{si2020adversarial} on NTU RGB+D 60 and the PKU-MMD I datasets. To compare with prior works, we report results when using 1\%, 5\%, 10\% and 20\% of the training data with labels for NTU RGB+60 and when using 1\% and 10\% of the labels for PKU-MMD I. The rest of the training set is used as the unlabeled data. 
 
The results in Table \ref{table:2} reveal that our method outperforms all previous methods on each benchmark.
We also demonstrate a large improvement over supervised only training, \ie~ training with only the available labeled data from randomly initialized weights. From these results we can see that our inter-skeleton contrastive learning is especially suited to learn from both unlabeled and labeled skeleton data in order to boost the performance of 3D action recognition. 

\noindent\textbf{Transfer Learning for 3D Action Recognition.} To evaluate 
if knowledge gained from a source dataset generalizes to a different target dataset, we also consider transfer learning. In this setting, an encoder network is first trained on the source dataset for our inter-skeleton contrastive task, followed by jointly finetuning the pretrained encoder and a classifier on a target dataset for action recognition. As in Lin \etal~\cite{ms2l}, we use NTU RGB+D 60 and PKU-MMD I as the source datasets and PKU-MMD II as the target dataset.  
Table \ref{table:4} shows our features are just as or more transferable than those of Zheng \etal~\cite{longtan} and Lin \etal~\cite{ms2l}, especially for transfer from PKU-MMD I to PKU-MMD II which are from same domain. 
Thus, the knowledge gained by our method from a source dataset can improve action classification accuracy on a different target set, especially one with a similar domain.

\begin{table}[t!] 
 \centering
  \resizebox{0.87\linewidth}{!}{
 \begin{tabular}{lccc}
 \toprule
& \multicolumn{2}{c}{\textbf{Transfer to PKU-MMD II}} \\
\cmidrule(lr){2-3}
& PKU-MMD I &  NTU RGB+D 60 \\
  \midrule 
 Zheng \etal~\cite{longtan}  & 43.6 & 44.8    \\
 Lin \etal~\cite{ms2l}       & 44.1 & \textbf{45.8}    \\
\textbf{\textit{This paper}} & \textbf{45.1} & \textbf{45.9 }   \\
 \bottomrule
 \end{tabular}
 }
 \caption{\textbf{Transfer learning for 3D action recognition.} 
 All results by Zheng~\etal ~provided by Lin \etal ~in \cite{ms2l}. Knowledge gained via inter-skeleton contrastive pretraining transfers well, especially when source and target datasets are more similar.  
} \label{table:4}
\vspace{-1em}
\end{table}
\subsection{Ablation Studies}
\label{sec:ablation}
We now ablate the effect of each of our skeleton augmentations and demonstrate the effectiveness of our inter-skeleton contrastive learning. These ablations are performed on the cross-view protocol of NTU RGB+D 60 for the downstream task of 3D action recognition. As before, after pre-training the models with our contrastive self-supervision methods, we train a linear classifier with action labels on top of the frozen features of the query encoder $f_{q}$. 

\noindent\textbf{Benefit of Skeleton Augmentation.} First, we show the benefit of each of the proposed skeleton augmentations when learning from a single input skeleton representation. We choose as skeleton augmentation function $D$, either pose augmentation, joint jitter, temporal crop-resize or combinations thereof, and train an intra-skeleton contrastive model as described in Section~\ref{section-3.2}. 

Table \ref{table:5} shows the accuracy of our augmentations with each input representation.
We find that all of the proposed spatial and temporal skeleton augmentations individually perform better than using no augmentation. Thereby, reinforcing our claim that learning invariances to spatial changes like viewpoints, scale and joint perturbations, or, temporal changes such as delay and speed result in learning good action features.
The composition of augmentations further improves the accuracy by a considerable margin for all input representations, with the best combination being the inclusion of all three augmentation functions. For example, the final accuracy with the $X^{\textit{IMG}}$ representation is a $\sim$10\% increase over using only pose augmentation and $\sim$28\% over using no augmentation. 

\begin{table}[t!] 
 \centering
 \resizebox{\linewidth}{!}{
 \begin{tabular}{lllcccc}
 \toprule
  \multicolumn{3}{c}{\textbf{Augmentations}} &  \multicolumn{3}{c}{\textbf{Downstream Representation}} \\ 
  \cmidrule(lr){1-3}\cmidrule(lr){4-6} 
  Temporal & Pose & Joint & \multirow{2}{*}{$X^{\textit{IMG}}$} &\multirow{2}{*}{$X^{\textit{STG}}$} &\multirow{2}{*}{$X^{\textit{SEQ}}$} \\\addlinespace[-0.4ex]
 Crop-resize & & Jitter &   &  &   \\
  \midrule
  - & - & - &  51.0    &  51.4  &     50.0 \\
 \checkmark & - &  -  &  62.5    &  53.5  &     64.1 \\
 -  & \checkmark & - & 69.8    &  63.8  &     71.7 \\
 - & - & \checkmark    &  74.6    &  66.1  &     75.2 \\
 \checkmark & \checkmark & -                 &  73.2    &  69.3  &     73.8 \\
\checkmark & - & \checkmark                   &  77.0    &  68.3  &     80.0 \\
 \checkmark & \checkmark & \checkmark    &\textbf{79.6}   &\textbf{72.5 }&   \textbf{82.5}  \\
      
 \bottomrule
 \end{tabular}
 }
 \caption{\textbf{Benefit of skeleton augmentation}. We ablate the effect of our augmentations with 3D action recognition on NTU RGB+D 60. Combining all three augmentations generates strong positive pairs for increased accuracy, no matter the 3D action representation.} 
 \label{table:5}
 \vspace{-1em}
\end{table}

\begin{figure}[t!]
\centering 
\includegraphics[width=\linewidth]{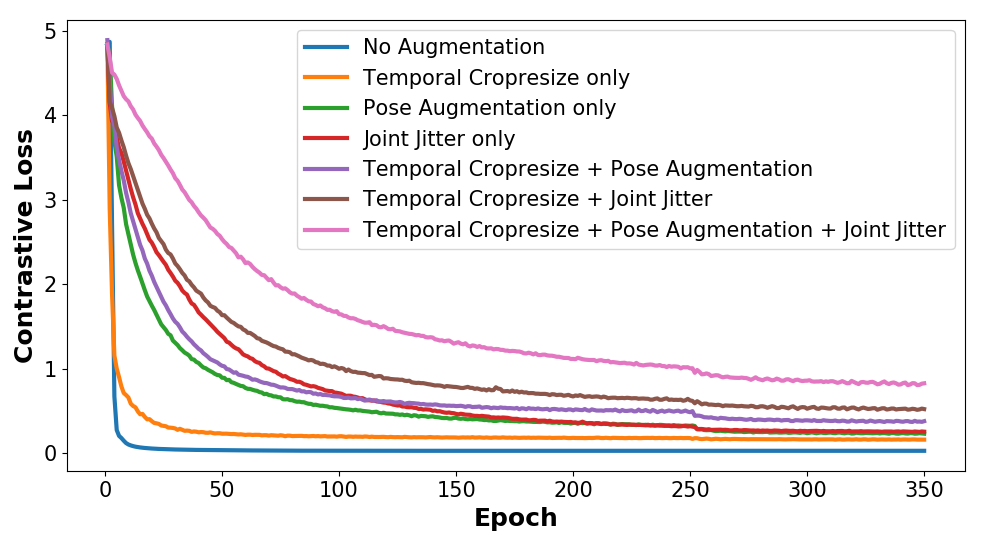}
\vspace{-1em}
\caption{
\textbf{Skeleton augmentation loss  curves}. Our proposed spatial and temporal skeleton augmentations make the contrastive task more difficult which prevents early saturation of the loss. The network is forced to focus more on commonalities in pose and joint motion dynamics to learn the similarities.
}
\label{fig:6}
\vspace{-0.5em}
\end{figure}

The benefit of our proposed skeleton augmentations are also reflected in the contrastive pre-training plots in Figure~\ref{fig:6}, which demonstrate that without augmentation the contrastive task is too easy, resulting in early saturation of the loss and poor features. With our spatial and temporal augmentations the contrastive task becomes more difficult as the network is encouraged to focus more on the pose and spatio-temporal movements of the joints, thereby improving downstream accuracy. Thus the combination of all our augmentations result in learning our best 3D action features.

\begin{table}[t!] 
 \centering
  \resizebox{\linewidth}{!}{
 \begin{tabular}{lccc}
 \toprule
  & \multicolumn{3}{c}{\textbf{Downstream Representation}} \\ 
  \cmidrule(lr){2-4}
  \textbf{Pretraining} &  {$X^{IMG}$} & ${X^{STG}}$ & ${X^{SEQ}}$\\
\midrule
 Intra ({$X^{{IMG}}$} only) & 79.6 & -  & - \\
 Intra (${X^{STG}}$ only) & - & 72.5  & - \\
 Intra (${X^{SEQ}}$ only) & - & - & 82.5 \\
 \midrule
 Inter ({$X^{{IMG}}$}, ${X^{STG}}$) & 80.0  & 78.0 & - \\
 Inter ({$X^{{IMG}}$}, ${X^{SEQ}}$) & \textbf{81.7} & - & 83.0\\
 Inter ({$X^{{SEQ}}$}, ${X^{STG}}$) & - &  78.9 & \textbf{85.2} \\
 Inter ({$X^{{IMG}}$}, {$X^{{SEQ}}$}, ${X^{STG}}$) &\textbf{81.2} &  \textbf{81.6} & \textbf{85.4}\\
 \bottomrule
 \end{tabular}
 }
 \caption{\textbf{Intra-skeleton \textit{vs.} Inter-skeleton}. Training alongside a second input representation in our inter-skeleton contrast results in better features for all input representations, regardless of the pair used. Note that a representation can only be used in the downstream task when it is present in pre-training. Ablation performed on 3D action recognition with NTU RGB+D 60.  
} \label{table:6}
\vspace{-2em}
\end{table} 

\noindent\textbf{Intra-Skeleton \textit{vs.} Inter-Skeleton.} Next, we examine the effectiveness of learning two skeleton representations together in our inter-skeleton framework over learning from each input representation separately (intra-skeleton). 
While our inter-skeleton network pre-trains two input skeleton representations alongside one another, to allow for fair comparison to the intra-skeleton network we train and test the downstream action recognition model with each input representation separately. The results of combining multiple representations in downstream tasks are presented in supplementary.

Table \ref{table:6} shows the accuracy of our inter-skeleton contrast  compared to the intra-skeleton baseline for each skeleton representation. We first observe that pre-training with any two skeleton representations side by side in our inter-skeleton contrast is considerably better than only learning with a single representation as in the intra-skeleton contrast. For example, the accuracy with $X^{STG}$ increases by 6\% when pre-trained together with $X^{SEQ}$ in our inter-skeleton contrast model. A similar increase of 5\% occurs when pre-training alongside $X^{IMG}$. We find this to be the case with each skeleton representation; regardless of the second representation it is trained alongside in the inter-skeleton contrast, there is an increase in performance. We also tried training all three skeleton representations together. While this does give the best result, the improvement is outweighed by the computational cost of training all three representations simultaneously. 
Overall, these results reinforce our claim that learning invariance to skeleton augmentations alone leads to sub-optimal features and learning additional invariance to  skeleton-representations results in a better feature space.

\section{Conclusion}
In this work, we presented a method for self-supervised learning of 3D skeleton data. We design a contrastive learning framework that relies on novel skeleton augmentations and multiple skeleton-representations to learn spatio-temporal dynamics of the skeleton sequences. Our comprehensive evaluation with different skeleton augmentations and skeleton-representation pairs reveal that learning invariance to our spatio-temporal augmentations and contrasting sequence-based and graph-based representations with each other results in best action features. The final model achieves considerable performance gains and outperforms prior state-of-the-art in self-supervised learning for multiple downstream tasks on NTU RGB+D 60 \& 120 and PKU-MMD.
\begin{acks}
This work is part of the research programme Perspectief EDL with project number P16-25 project 3, which is financed by the Dutch Research Council (NWO) domain Applied and Engineering/ Sciences (TTW).
\end{acks}

\bibliographystyle{ACM-Reference-Format}
\balance
\bibliography{sample-base}

\clearpage 
\appendix
\section{ Appendix}
In this Appendix, we provide details on the training procedure for each downstream task in Section~\ref{sec:downstream_imp} and a comparison of our method to supervised-approaches for skeleton-based action recognition in Section~\ref{sec:supervised_comp}. We examine the effect of the hyperparameters of our proposed augmentations in Section~\ref{sec:aug_ablations}. Finally, we show the performance of combining multiple-skeleton representations  for the downstream task of action recognition in Section~\ref{multi} and  provide some qualitative results of our method in Section~\ref{sec:qual_res}.

\subsection{Downstream Training  Details}
\label{sec:downstream_imp}
For the downstream tasks we follow Chen \etal~\cite{simclr} and remove the projection head of the pre-trained query encoder, as the projection head tends to focus mostly on information specific to the pretext task. For the 3D action recognition tasks, we then append a classifier to the pre-trained query encoder, while for 3D action retrieval we directly use the feature space without adding a classification head. The dimensionality of the feature space is dependent on the input skeleton-representation used in the downstream task. It is either 4096 (for $X^{\textit{IMG}}$), 2048 (for $X^{\textit{SEQ}}$) or 256 (for $X^{\textit{STG}}$). For downstream tasks we use a temporal crop of length 64. During training this is sampled randomly, while for evaluation we sample a center crop. 

\noindent\textbf{3D Action Recognition.} For this task,  the weights of the pre-trained encoder are frozen and only the linear classifier is trained as in \cite{ms2l,Unsup3DPose}. An SGD optimizer is used with a momentum of 0.9 and learning rate of 0.1. The linear classifier is trained for a total of 80 epochs and learning rate is reduced by a factor of 10 after the 50th and 70th epoch. 

\noindent\textbf{3D Action Retrieval} For this task we follow ~\cite{pc} to extract the  encoder features of the training set. Then, we apply  a $k$NN classifier with $k{=}1$ using these features and their corresponding action labels  to assign action classes. Finally, during testing we assign to the unseen sample the action class of the closest neighbour in the training set.  

\noindent\textbf{Semi-Supervised 3D Action Recognition.} For this task, we finetune both the classifier and the pre-trained encoder weights jointly as in \cite{ms2l}.  An Adam optimizer is used to train the network  for a total of 50 epochs with a learning rate of 0.0001,  which is reduced by a factor of 10 after both the 30th and 40th epoch. 

\noindent\textbf{Transfer Learning for 3D Action Recognition.} For this task we again follow \cite{ms2l} and finetune the classifier and the pre-trained encoder together. An Adam optimizer is used to train the network  for a total of 50 epochs with a learning rate of 0.0001  which is reduced by a factor of 10 after 30th and 40th epoch. 

\subsection{Supervised Approaches}
\label{sec:supervised_comp}
While our method outperforms prior self-supervised learning works for 3D action recognition, it is also useful to know how this compares to state-of-the-art supervised approaches. Table \ref{table:7} shows the performance of various supervised approaches on the NTU 60 \& 120 datasets.  We compare these results to the performance of our sequence-based query encoder $f^{\textit{SEQ}}$ (a simple 3-layer Bi-GRU)  trained end-to-end from randomly initialized weights (supervised-only) and finetuned end-to-end from the weights learnt from our inter-skeleton contrastive learning approach (with pre-training). Note that this setting is different to the experiment performed in the main paper, which only finetunes the final layer in order to demonstrate the raw performance of the features, rather than the boost they can provide to supervised training.
It is evident from the table our method is competitive with many supervised approaches, even though the encoder we use is not state-of-the-art. It is also clear that our contrastive  pre-training can boost the performance over supervised-only training. It is likely our inter-skeleton contrastive pre-training can also be used to boost the performance of more complex state-of-the-art encoders too.

\begin{table}[h!] 
 \centering
 \resizebox{\linewidth}{!}{
 \begin{tabular}{lcccc}
 \toprule
  & \multicolumn{2}{c}{\textbf{NTU RGB+D 60}} & \multicolumn{2}{c}{\textbf{NTU RGB+D 120}} \\
  \cmidrule(lr){2-3} \cmidrule(lr){4-5} 
   \textbf{Method} &  x-view & x-sub & x-setup & x-sub \\ 
  \midrule
   PA-LSTM~\cite{Shahroudy_2016_NTURGBD}  &  52.8  & 50.1 & 26.3  & 25.5       \\
   ST-LSTM~\cite{rnn1}  &  77.7  & 69.2 & 57.9  & 55.7       \\
   GCA-LSTM~\cite{rnn4}  &  84.0  & 76.1  & 59.2  & 58.3      \\
   VA-LSTM~\cite{rnn3}  &  87.7  & 79.4 & -  & -     \\
   ST-GCN ~\cite{stgcn2018aaai}  &  88.3  & 81.5 & 73.2  & 70.7 \\
   Shift-GCN ~\cite{gcn4}  &  96.5	& 90.7 & 85.9	& 87.6  \\
   MS-G3D Net ~\cite{gcn5}  &   96.2 & 91.5 & 86.9 & 88.4 \\
 \midrule
\textit{\textbf{This paper}} (supervised-only)  &  {87.8}  & {72.9} & {68.2}  & {66.3} \\
\textit{\textbf{This paper}} (with-pretraining)  &  {90.4}  & {79.3} & {75.4}  & {73.1} \\
 \bottomrule
 \end{tabular}
 }
 \caption{
 \textbf{Comparison with supervised only training}{ for 3D action recognition.
 Pre-training with our inter-skeleton contrast improves the performance over supervised only training, especially for the more challenging cross-subject and cross-setup protocols. 
 }}
\label{table:7}
\end{table}

\subsection{Augmentation Hyperparameter Ablations}
\label{sec:aug_ablations}
 In this section we study the impact of hyperparameters $|j|$ and $L_{\textit{ratio}}$  of the spatial joint jittering and temporal crop-resize augmentations on the downstream performance. We use $X^{IMG}$ skeleton representation  and  evaluate on the cross-view protocol of NTU RGB+D 60 for the downstream task of 3D action classification.  We first pre-train an  intra-contrastive framework using $X^{IMG}$ representation with only the relevant augmentation and then  train a linear classifier with action labels on top of the frozen features of the query encoder $f_{q}$. 

\subsubsection{Effect of number joints to jitter $|j|$}

Here, we ablate over the number of joints to jitter $|j|$ in our joint jittering augmentation. This parameter controls the number  of joints to be jittered for the augmented view. Table ~\ref{table:8} shows the downstream 3D action classification performance of  different values of  $|j|$. 
We found that jittering around half the joints ($|j| {=} 10, 15$) performed best. Using very small or a large values for $|j|$ \eg~ 2 or 20 is sub-optimal as with too few jittered joints the augmented views become highly similar, while with many jittered joints there remains little commonality between the augmented sequences.
For all our experiments we use $|j| {=} 15 $  in our joint jittering augmentation as it achieve best downstream performance. 
\begin{table}[h!] 
 \centering
  \resizebox{0.85\linewidth}{!}{
 \begin{tabular}{lcccccc}
 \toprule
& \multicolumn{5}{c}{\textbf{Number of jittered joints  $|j|$}} \\
  \cmidrule(lr){2-6}
\textbf{Augmentation} & $2$ & $5$ &  $10$ & $15$ & $20$ \\
  \midrule 
 Spatial-Jittering & 65.6 & 67.5 & 69.4 & \textbf{74.6} & 70.6    \\
 \bottomrule
 \end{tabular}
 }
 \caption{\textbf{Effect of number of joints to jitter} on the downstream task  of 3D action classification on cross-view protocol of NTU RGB+D 60. Increasing the number of joints to jitter improves the downstream performance.  
} \label{table:8}
\end{table}

\subsubsection{Effect of temporal length ratio $L_{\textit{ratio}}$ }
We next ablate over the distribution from which temporal length ratio ${L}_{\textit{ratio}} \in [l_{\textit{min}}, 1.0]$ is sampled in our temporal crop-resize augmentation, see Equation~\eqref{eq:cropresize}.  The parameter $l_{\textit{min}}$ controls the minimum length of the temporal crop, which can be sampled for the augmented view. Table ~\ref{table:9} shows the 3D action classification performance with different minimum samples lengths ${l}_{\textit{min}}$. A smaller $l_{\textit{min}}$, and thus a larger temporal range improves the downstream performance. We therefore use $l_{\textit{min}} {=} 0.1$, \ie. ${L}_{\textit{ratio}} \in [0.1, 1.0]$, for all our experiments.

\begin{table}[h!] 
 \centering
  \resizebox{0.7\linewidth}{!}{
 \begin{tabular}{lcccc}
 \toprule
& \multicolumn{3}{c}{$l_{min}$} \\
 \cmidrule(lr){2-4}
\textbf{Augmentation} &  $0.1$ & $0.3$ & $0.5$ \\
  \midrule 
 Temporal Crop-Resize& \textbf{62.5} & 62.0 & 60.8 \\
 \bottomrule
 \end{tabular}
 }
 \caption{\textbf{Effect of temporal length ratio} on the downstream task  of 3D action classification on cross-view protocol of NTU RGB+D 60. The bigger the range, the better the downstream performance.  
} \label{table:9}
\end{table}

\subsection{Multi-representation Downstream}
\label{multi}
In this section, we examine the effect of combining skeleton representations when finetuning for the downstream task of 3D action recognition. All of our previous results use only one representation in the downstream task for efficiency, even when representations are trained together in our inter-skeleton contrast. Here we report the results of combining representations in the downstream task for both intra and inter-skeleton contrast. For intra-skeleton, each skeleton representation is first pretrained separately (see Section~\ref{section-3.2}) and then their query encoders are combined for the downstream task. For inter-skeleton two skeleton representations are pretrained together (see Section~\ref{section:-3.3}) with their query encoders also combined for the downstream task. Table~\ref{table:10} shows the results of these experiments alongside the results when using only one representation during the downstream task from Table~\ref{table:6}. We again evaluate on the cross-view protocol of NTU RGB+D 60 by training a linear classifier on frozen features. In Table~\ref{table:10} we also highlight the number of parameters needed for each representation in this downstream task. The downstream encoders (i.e query encoders) for the skeleton representations are as a 3-Layer BI-GRU with $H{=}$1024 units)  for {$X^{{SEQ}}$}, an HCN~\cite{HCN} model for {$X^{{IMG}}$} and   a joint-based  A-GCN \cite{2sagcn2019cvpr} network for {$X^{{STG}}$} (see Implementation details Section~\ref{section:implementation}).

From Table~\ref{table:10}, we first observe when combining representations in the downstream task pretraining with inter-skeleton contrast outperforms the intra-skeleton pretraining for all combinations. In this setting both the computational costs required for pretraining and inference  of the intra and inter-skeleton contrast are same as training two representations separately requires the same computation as training them together (inter), thereby showing the superiority of our inter-skeleton contrast.

As we saw in the main paper, inter-skeleton contrast shows considerable improvement in performance over the intra-skeleton contrast for the each single representation downstream evaluation, with {$X^{{SEQ}}$} obtaining the best results. However, it is worth noting that while the number of parameters required for inference are the same, the inter-skeleton does require additional computational resources for pretraining since each representation is required to be pre-trained with one of the other skeleton representations while in intra-skeleton each representation is pretrained alone.

We also observe that combining representations in the downstream task improves over using a single representation in the majority of cases,  with the combination  {$X^{{SEQ}}$} and {$X^{{STG}}$} showing the best results. Note that this improvement comes with an additional cost of model size during the inference time. With these results we can conclude that our model can be used for all skeleton representations individually or in combination based on the trade off between the pretraining computational cost, the inference model size and the performance.

\begin{table}[h!] 
 \centering
  \resizebox{\linewidth}{!}{
 \begin{tabular}{lccc}
 \toprule
  \textbf{Downstream Reps.} & {Intra} & {Inter} & \# Inference Params.  \\
\midrule
 {$X^{{IMG}}$} & 79.6 & 81.7 & 1.0M  \\
 {$X^{{STG}}$} & 72.5 & 78.9 & 3.0M  \\
 {$X^{{SEQ}}$} & 82.5 & \textbf{85.2} & 10.0M \\
 \midrule
 {$X^{{IMG}}$} + {$X^{{STG}}$} & 80.3 & 81.8 & 4.0M \\
 {$X^{{IMG}}$} + {$X^{{SEQ}}$} & 80.3 & 82.6 & 11.0M \\
 {$X^{{SEQ}}$} + {$X^{{STG}}$} & 84.5 & \textbf{86.0} & 14.0M \\
 
 \bottomrule
 \end{tabular}
 }
 \caption{\textbf{Combining representations for 3D action recognition.} We show the trade-off between accuracy and number of parameters involved in the downstream task when using two representations to fine-tune the features learnt from both intra and inter-skeleton pretraining. Pretraining with our inter-skeleton contrast learns better features for each representation whether used individually or combined.
} \label{table:10}
\vspace{-2em}
\end{table} 

\subsection{Qualitative Results}
\label{sec:qual_res}
\subsubsection{Visualization of learned features} 
\begin{figure*}[h]
\centering
\includegraphics[width=0.92\linewidth]{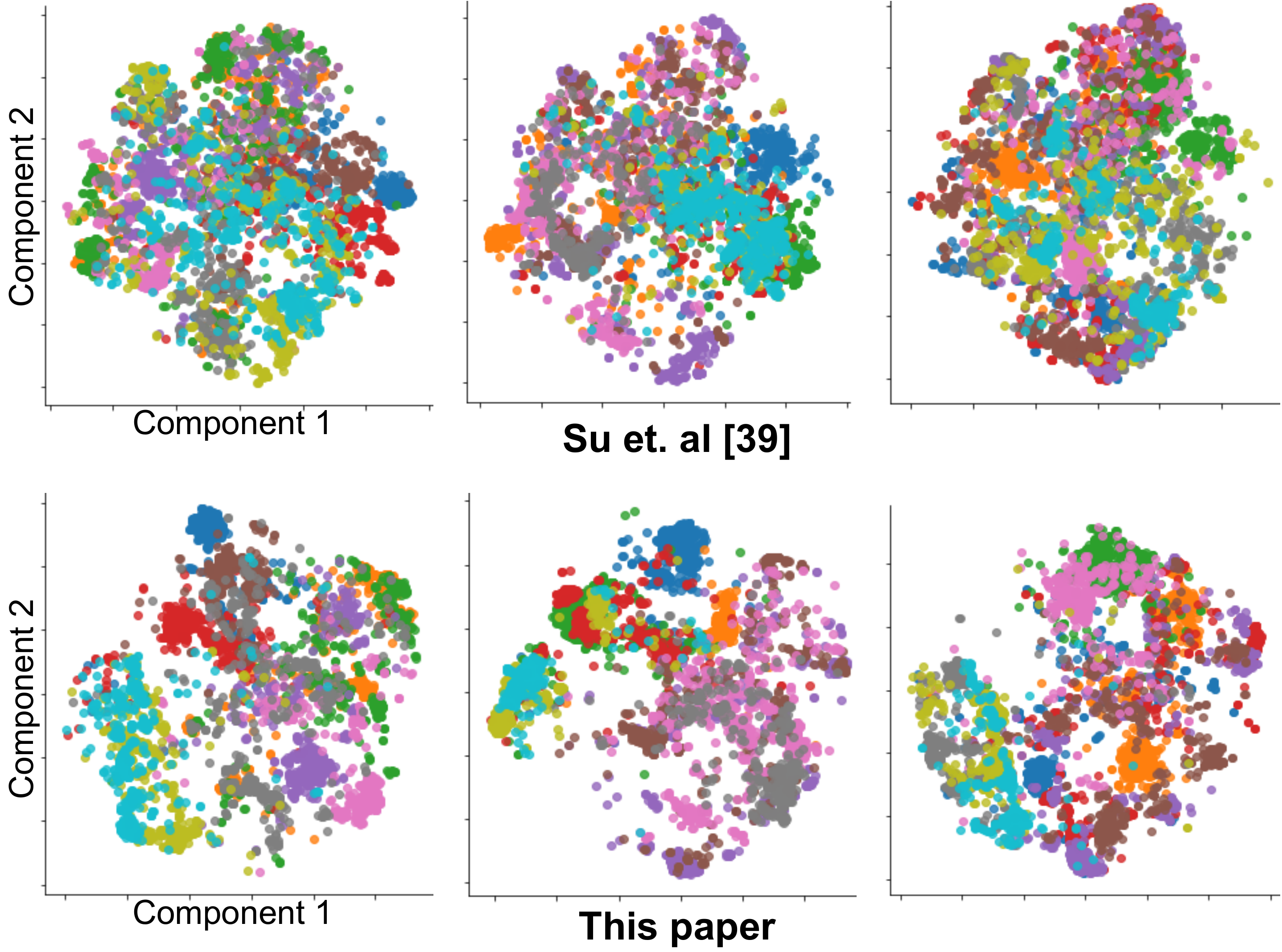}
\caption{
\textbf{t-SNE visualization } of learned features on NTU RGB+D 60 dataset. Each plot shows the features of 10 randomly selected action classes. Top row shows the features learned by Su \etal~\cite{pc}  and bottom row shows  the corresponding features learned by  our inter-skeleton contrastive learning. Our methods learns a more discriminatory feature space  forming better clusters  which are more dense with most samples from same action class and distant from other clusters as compared to \cite{pc}. 
}
\label{fig:7}
\end{figure*}

First we visualize the features by our inter-skeleton contrastive learning in comparison to those learned by Su \etal~\cite{pc}, one of the best performing methods on both the 3D action recognition and retrieval tasks. We randomly select 10 of the 60 action classes, so as not to overcrowd the figure, and plot their features using t-SNE. This is repeated three times for three different subsets of action classes. We observe from the Figure \ref{fig:7} that the features learned by our method form better clusters and are therefore more discriminatory and more suitable for the downstream tasks of action recognition and retrieval. 

\subsubsection{ 3D Action retrieval  results} 
\begin{figure*}[h]
\centering
\includegraphics[width=0.98\linewidth]{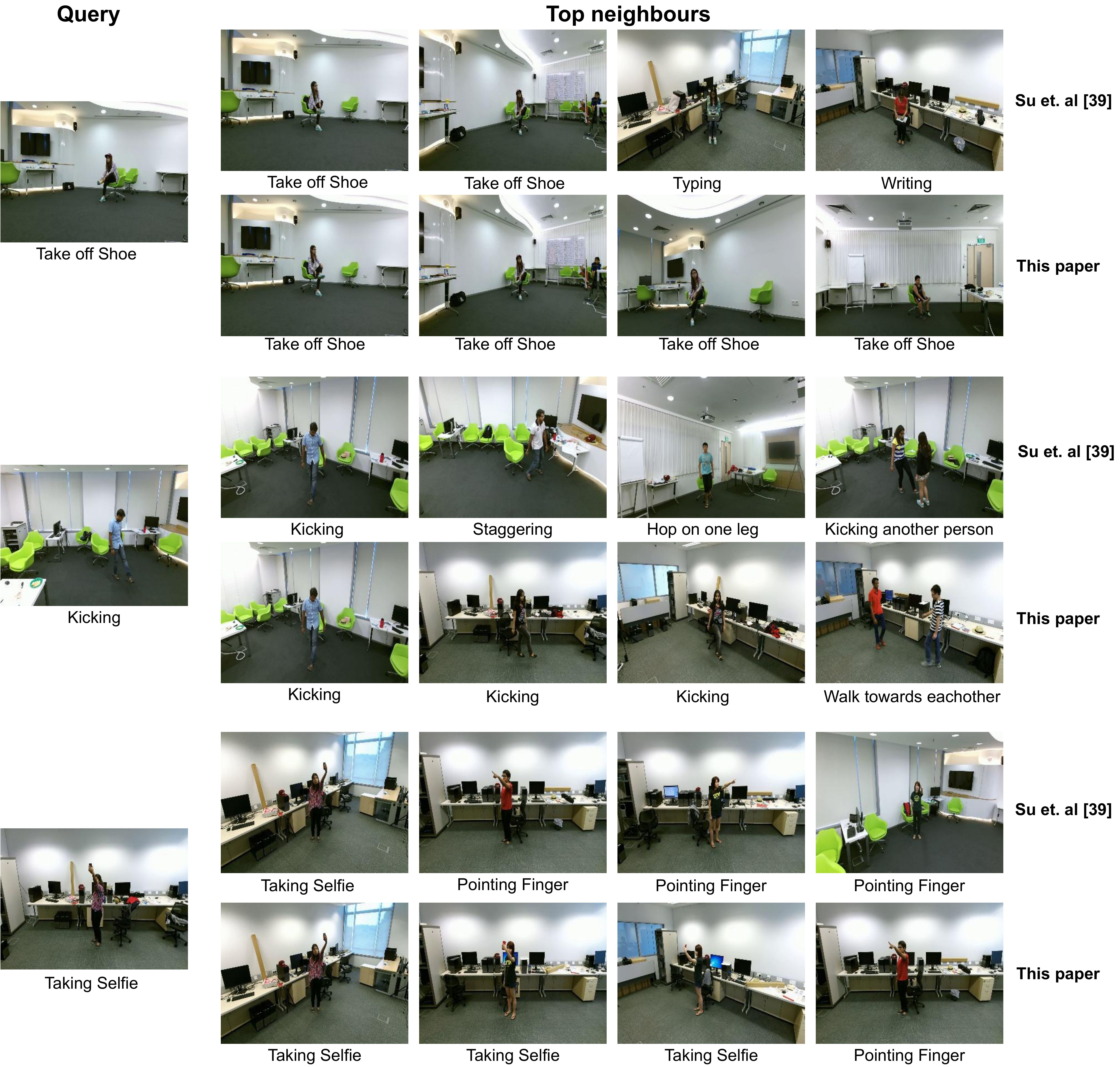}
\caption{
\textbf{3D Action retrieval results}  on NTU RGB+D 60 dataset. For each query, the first row shows nearest neighbours learned by Su \etal~\cite{pc} and the second row shows the nearest neighbours in the feature space learnt by our inter-skeleton contrastive learning. For our method most neighbours belong to the same action classes.  All results were obtained using 3D skeleton data, however, for the ease of visualization/interpretation we show the corresponding RGB videos instead of the skeleton sequences.
}
\label{fig:8}
\end{figure*}

In Figure~\ref{fig:8}, we visualize the results of 3D action retrieval. For a given query video we retrieve the top four nearest neighbours in the feature space learned by  Su \etal~\cite{pc} and by our inter-skeleton contrastive learning. We observe from Figure  \ref{fig:8} that the nearest neighbours in the feature space are generally more relevant to the query when using our method. The videos retrieved by Su~\etal~tend to be from different actions, which contain similar body poses. For instance `kicking', `staggering' and `hop on one leg' all contain poses with one leg off the floor. Instead, our method is able to better focus on the motion of the query action and retrieve other instances of the same action.

\end{document}